\documentclass[11pt]{article}
\usepackage{acl}

\usepackage[T2A, T1]{fontenc}
\usepackage[utf8]{inputenc}
\usepackage{times}

\usepackage[russian, czech, english]{babel}
\usepackage{CJKutf8}

\usepackage{latexsym}
\usepackage{float}
\usepackage{graphicx}
\DeclareGraphicsExtensions{.pdf,.png,.jpg,.jpeg}
\usepackage{pdflscape}
\usepackage{booktabs}
\usepackage{multirow}
\usepackage{longtable}
\usepackage{xcolor}
\usepackage{subcaption}
\usepackage{xspace}
\usepackage{placeins}

\usepackage[skins,breakable]{tcolorbox}
\tcbset{
  promptstyle/.style={
    enhanced jigsaw,
    breakable,
    boxrule=0.5pt,
    colframe=black,
    colback=white,
    arc=2pt,
    left=5pt,right=5pt,top=5pt,bottom=5pt,
    before skip=5pt,after skip=5pt,
    box align=top,halign=left,
    segmentation style={top}{draw=none},
    segmentation style={bottom}{draw=none},
  }
}
\newtcolorbox{promptquote}{promptstyle}

\usepackage{fvextra}
\DefineVerbatimEnvironment{verbatim}{Verbatim}{breaklines,breakanywhere}
\usepackage{algorithm}
\usepackage{algpseudocode}

\usepackage{hyperref}
\newcommand{\ayaeightb}{\texttt{Aya-Expanse-8B}\xspace}

\newcommand{\GPTSW}{\texttt{GPT-SW3-6.7B}\xspace}

\newcommand{\tri}{\texttt{Tri-7B}\xspace}

\newcommand{\glm}{\texttt{GLM4-9B}\xspace}

\newcommand{\phimodel}{\texttt{Phi4-mini-instruct}\xspace}

\newcommand{\towerplus}{\texttt{TowerPlus-9B}\xspace}

\newcommand{\llamainstruct}{\texttt{Llama3-8B-Instruct}\xspace}

\newcommand{\xtower}{\texttt{xTower}\xspace}

\newcommand{\gptfour}{\texttt{GPT4.1}\xspace}

\newcommand{\claude}{\texttt{Claude-4}\xspace}

\newcommand{\deepseek}{\texttt{DeepSeek-v3}\xspace}

\newcommand{\gemmatwo}{\texttt{Gemma2-27B}\xspace}

\newcommand{\llamathree}{\texttt{Llama3.3-70B}\xspace}

\newcommand{\PAIlogo}{\raisebox{3.4pt}{\includegraphics[scale=0.1]{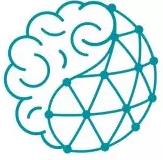}}}

\title{\textbf{Can QE-informed (Re)Translation lead to Error Correction?}}
\author{
Govardhan Padmanabhan\PAIlogo \\\\
\PAIlogo Institute for People-Centred AI \\
University of Surrey, United Kingdom \\
\texttt{gp00816@surrey.ac.uk}
}

\begin{document}
\maketitle

\begin{abstract}
The paper presents two approaches submitted to the WMT 2025 Automated Translation Quality Evaluation Systems Task 3 - Quality Estimation (QE)-informed Segment-level Error Correction. While jointly training QE systems with Automatic Post-Editing (APE) has shown improved performance for both tasks, APE systems are still known to \textit{overcorrect} the output of Machine Translation (MT), leading to a degradation in performance. We investigate a simple training-free approach - QE-informed Retranslation, and compare it with another within the same training-free paradigm. Our winning approach selects the highest-quality translation from multiple candidates generated by different LLMs. The second approach, more akin to APE, instructs an LLM to replace error substrings as specified in the provided QE explanation(s). A conditional heuristic was employed to minimise the number of edits, with the aim of maximising the Gain-to-Edit ratio. The two proposed approaches achieved a $\Delta$COMET score of $0.0201$ and $-0.0108$, respectively, leading the first approach to achieve the winning position on the subtask leaderboard.
\end{abstract}

\section{Introduction}
Large Language Models (LLMs) have advanced the field of Machine Translation (MT), given their support for longer input context length and ability to generate text in a natural tone. However, translation quality is still limited for languages other than English and for domain-specific translations~\cite{fernandes2025llmsunderstandtranslationsevaluating}. Evaluating MT output for quality is critical to understand the reliability and suitability of translation systems, and more importantly, to be able to perform accurate corrections. The WMT24 Metrics Shared Task found that neural-based learned metrics like COMET or xCOMET were superior when evaluating LLM-generated translations, compared to the traditional statistical metrics like BLEU, or chrF~\cite{freitag-etal-2024-llms}. 

As LLMs are typically trained on a large general dataset, performance in domain-specific MT can often fall short as they may not properly render key terminologies or stylistic conventions. As such, Automatic Post-Editing (APE) is vital to fixing MT errors. However, they are prone to overcorrecting. One such example of mitigating over-correcting, proposed by~\citet{doiGivingFresh}, is to utilize word-level Quality Estimation (QE) to limit edits only on the specified error segments. Despite recent efforts to reduce overcorrection~\cite{deoghare-etal-2023-quality,deoghare-etal-2024-together}, APE models still fall short of the required semantically coherent output. Therefore, we ask the titular question - ``Can QE-informed re-translation help overcome MT errors?'', and discuss two approaches as a comparative evaluation.\\

This paper describes two participation systems, both utilizing pre-trained and open-source LLMs, for the WMT 2025 Automated Translation Quality Evaluation Systems Task 3 - QE-informed Segment-level Error Correction. The primary approach leverages multiple LLMs for MT and selects the best output using QE. In the secondary approach, an LLM is prompted to replace error-segments in the provided MT. These error segments are identified by the explainable QE provided within the dataset.

\section{Related Work}

\begin{figure*}[h!]  
    \centering
    \includegraphics[width=0.9\linewidth]{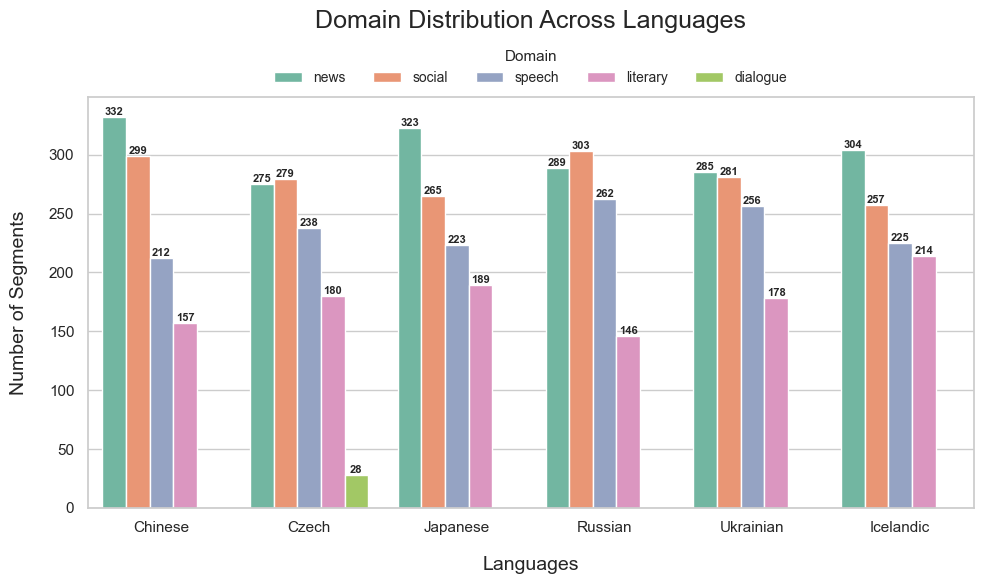}
    \caption{Domain distribution within each language}
    \label{fig:placeholder}
\end{figure*}

\paragraph{Quality Estimation (QE)} QE is an automated evaluation framework that predicts a score indicating whether the translation is good or not~\cite{10.1162/coli_r_00352}. COMET (Cross-Lingual Optimized Metric for Evaluation of Translation) is an automatic metric to evaluate the quality of machine translation using deep learning models~\cite{rei-etal-2020-comet}. COMETKiwi is a hybrid machine translation quality estimation (MTQE) model that combines COMET and OpenKiwi~\cite{rei-etal-2022-cometkiwi}. It achieved top performance in the WMT 2022 shared task and has since been widely adopted as a state-of-the-art benchmark in MTQE.

\paragraph{Automatic Post-Editing (APE)} APE is an automated system that corrects MT output, without human involvement~\cite{doCarmo2021review}.~\citet{chatterjee-etal-2018-combining} describe combining QE and APE in three ways: using QE as an APE activator when the MT output is poor, as guidance to help the APE decoder decide which tokens to change, and as a selector to choose between the raw MT and post-edited output.

\paragraph{WMT24 QE-APE} The previous year's WMT competition focused on sentence-level quality estimation and error span predictions~\cite{zerva-etal-2024-findings}. QE was further incorporated into APE. The dataset for the QE-APE primarily consisted of English–Hindi (En-Hi) and English–Tamil (En-Ta) pairs. The source (SRC) English sentence, the target (TGT) translation provided by an unspecified neural machine translator, and a human post-edit (PE) version of the translation made by native speakers were included to support both quality estimation and automatic post-editing tasks.

The HW-TSC team~\cite{yu-etal-2024-hw} utilized \llamainstruct for En-Hi pairs. The model first underwent continual pretraining using low-rank adaptation (LoRA) on SRC and TGT data, and was further supervised fine-tuned on the PE data with a custom prompt.  For En-Ta pairs, they trained a custom transformer then performed APE fine-tuning. This system achieved $0.851$ and $0.918$ COMET scores for En-Hi and En-Ta translation tasks, respectively.

The IT-Unbabel team utilized \xtower for generating corrected translations, along with a quality estimation model to decide whether to use the original translation or the \xtower output. This approach achieved $0.8646$ COMET score for En-Hi pairs, and $0.9163$ for En-Ta pairs.

\paragraph{} IT-Unbabel's solution of utilizing an LLM along with QE as a selector served as the inspiration for the primary approach

\section{Methodology}

\begin{figure*}[t!]
    \centering
    \includegraphics[width=0.89\linewidth]{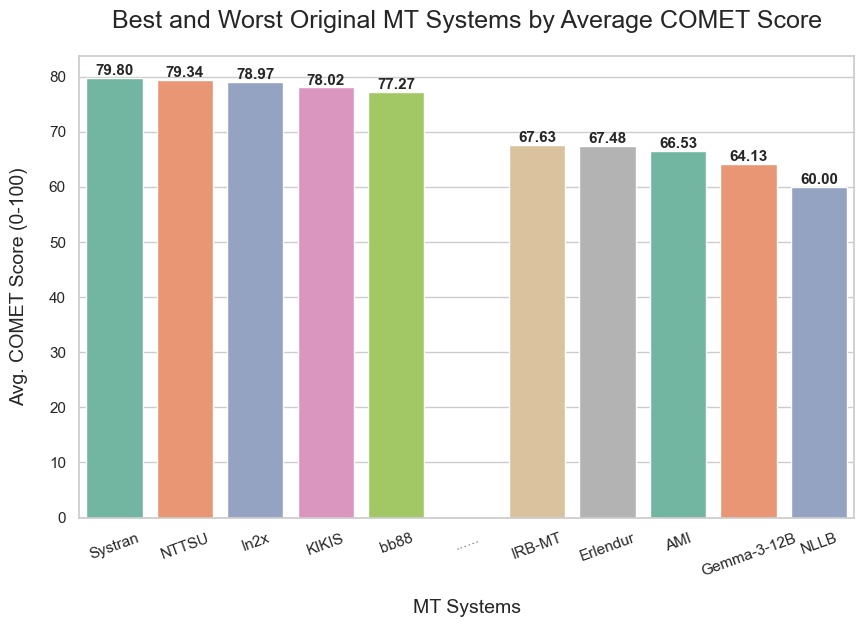}
    \caption{System performance in hypothesis\_segment}
    \label{fig:fig1score}
\end{figure*}

\subsection{Dataset Analysis}

The complete test data provided with the task was exclusively used. This dataset consists of $6,000$ machine translations from English to Chinese, Czech, Japanese, Icelandic, Russian, and Ukrainian, with $1,000$ instances for each language pair. The texts cover a range of domains, specifically news, social, speech, literary, and dialogue. As shown in Figure~\ref{fig:placeholder}, the domains are not evenly distributed. The \textit{news} domain is the most prominent overall with $1,808$ entries, while the \textit{dialogue} domain has the lowest representation with $28$ segments, all appearing only in the English-to-Czech group.

The original translations under the ``hypothesis\_segment'' column were generated by $38$ different translation systems, including LLMs like \gptfour, \claude, \deepseek, and more. Figure~\ref{fig:fig1score} shows the five best and worst systems according to their average translation quality scores.

\subsection{Approaches}

\subsubsection{Primary Approach - ``Best MT Wins''}

In this approach, multiple LLMs were used to translate the English texts from scratch, without additional information or context. The resulting candidate translations were then evaluated with the \textit{wmt22-cometkiwi-da} model, which provided QE scores based on the source English text and translated system outputs. The translation with the highest QE score was selected as the final output. By re-framing the APE task as re-translation, QE serves as a selector in place of traditional decoders. Similar approaches have also been explored, where QE is used in multi-hypothesis selection~\cite{yu-etal-2024-hw, 10.1007/978-3-319-77116-8_32, 10.1007/978-3-030-32233-5_28}, further supporting the view of QE as a decision mechanism in MT improvement. This approach therefore functions as a systematic probe of QE-based re-ranking, illustrating both its potential to establish an empirical upper bound on metric-based performance and its limitations in terms of computational cost and efficiency.

\ayaeightb, \GPTSW, \tri, \glm, \phimodel, and \towerplus models were used. These models were selected because of their reported performance, robustness, popularity, and recency. Models like \GPTSW, \tri, and \glm were selected owing to their specialized training in certain languages, specifically Icelandic, Japanese, and Chinese respectively.

The prompts for \tri, \phimodel, and \towerplus are reminiscent of their translation prompts available in their Huggingface model cards, being variations of: \begin{promptquote} Translate from English to \textit{\{Language\}}\end{promptquote}

System prompts for the other models were more involved, as using the same prompt resulted in some hallucinations, chain of thought, or worse translation quality during limited internal testing.

For \ayaeightb:
\begin{promptquote}
You are a helpful bilingual assistant that correctly translates the user's input text from English to \textit{\{Language\}}.

When translating, you must use the same tone and intent of the English text. You will include any and all special characters from the input.

If there is no proper translation for an English word or phrase, you can use the English word or phrase in place.
\end{promptquote}

\hfill\break

For \GPTSW:
\begin{promptquote}
<|endoftext|><s> \\
System: \\
You are a bilingual assistant that objectively translates the user's input text from English to Icelandic. \\
You will include any and all special characters from the input. \\
If there is no proper translation for an English word or phrase, you can use the English word or phrase in place. \\
<s> \\
User: \\
\textit{\{original English text\}} \\
<s> \\
bot:
\end{promptquote}

\hfill\break

For \glm:
\begin{promptquote}
    You are a bilingual assistant that correctly translates the user's input text from English to Chinese.
    
    When translating, you must use the same tone and intent of the English text. You will include any and all special characters from the input.
    
    If there is no proper translation for an English word or phrase, you can use the English word or phrase in place.
    
    Output only the translation!
\end{promptquote}

\hfill\break

Except \towerplus, the remaining models do not support all required languages. Hence, they translated only their supported language(s), and the rest were omitted. After all models were successfully executed, the QE score via COMETKiwi between all translations, including the original systems', was used to determine which MT to use.

\subsubsection{ Secondary Approach - ``Fill in the Blanks''}

The provided test data includes \textit{error spans} for the translations. Using fine-grained QE signals to guide targeted corrections, this approach investigates whether restricting edits to QE-highlighted segments could yield improvements with fewer changes, thereby increasing both efficiency and interpretability compared to full re-translation.

Using these error spans, the corresponding substring(s) in translation is replaced with a "\textbf{\_\_BLANK\_\_}" token, emulating a multilingual masked language modeling task. The text domain is also used to provide additional context. An example is included in the prompt to guide the model's behavior, serving as a one-shot example. Different examples were used in the prompt for different languages.

This approach utilizes \towerplus, an open source LLM with 9 billion parameters based on Gemma2. This model was selected because of its training for translation-related tasks along with instruction tuning, a context window of $8192$ tokens, and has support of the languages of this task: English, Chinese, Czech, Japanese, Icelandic, Russian, Ukrainian. It has also been shown to outperform larger parameter models like \gemmatwo and \llamathree on translation performance~\cite{rei2025towerplus}. The 9B variant was specifically selected due to resource and time constraints.

Provided below is the system prompt template for Russian language translations: 

\begin{promptquote}
You are a helpful assistant that corrects a Russian translation by filling in the blanks. Use the English sentence for context. Complete the task while maintaining the tone of a \textit{\{domain\}}. \\
Important - Do not use any of the specified wrong words. Replace each \_\_BLANK\_\_ token with an appropriate word or phrase that matches the original meaning, tone, and context of the English sentence.\newline

Example (social domain):\\
Russian with \_\_BLANK\_\_: 
\selectlanguage{russian} Многие молодые люди сегодня предпочитают \_\_BLANK\_\_ в кофейнях, а не дома.\\
\selectlanguage{english}English: Many young people today prefer to hang out in coffee shops rather than at home.\\
Wrong translation: 
\selectlanguage{russian}Многие молодые люди сегодня предпочитают учиться в кофейнях, а не дома.\\

\selectlanguage{english}Wrong words: ['
\selectlanguage{russian}учиться']\\
\selectlanguage{english}Corrected Russian sentence: 
\selectlanguage{russian}Многие молодыe люди сегодня предпочитают проводить время в кофейнях, а не дома.\newline

\selectlanguage{english}
Russian with \_\_BLANK\_\_: \textit{\{mt with \_\_BLANK\_\_\}}\\
English: \textit{\{source text\}}\\
Wrong translation: \textit{\{mt text\}} \\
Wrong words: \textit{\{list of substrings removed\}}\\
Corrected Russian sentence:\\
\end{promptquote}

To increase translation quality while keeping changes minimal (gain-to-edit ratio), a conditional masking heuristic based on error severity and overall QE score was employed. The pseudocode for this method is provided in Algorithm 1.

\begin{algorithm}[h!]
\caption{Conditional masking based on severity and score}
\begin{algorithmic}[1] 
\State \textbf{Original QE Score:} Float $x$
\If{$x >= 0.90$}
    \State Proceed without masking
\ElsIf{$x > 0.50$}
    \If{only minor severity error spans}
        \State Mask minor error spans
    \Else
        \State Mask all non-minor error spans
    \EndIf
\Else
    \State Mask all error spans
\EndIf
\end{algorithmic}
\end{algorithm}

\section{Results and Discussion}

Tables 1 and 3 present language-wise results for the primary and secondary approaches, respectively. The WMT'25 shared task tracks two main metrics: $\Delta$COMET and Gain-to-Edit Ratio (referred to as "G2E Ratio"), with the overall $\Delta$COMET score serving as the primary selection criterion. Here, $\Delta$COMET is computed as the difference between the COMET scores of the proposed approach and the baseline translation. The G2E Ratio is calculated as $\Delta$COMET divided by the total edit rate. BLEU and chrF++ scores are also included in the tables as supplementary information, but they are not the focus of the analysis. The best-performing $\Delta$COMET and G2E Ratio scores across the two tables are highlighted in bold.

\subsection{Primary Approach - ``Best MT Wins''}

\begin{table}[h!]
\centering
\resizebox{\columnwidth}{!}{%
\begin{tabular}{@{}ccccc@{}}
\toprule
Language & $\Delta$COMET & G2E Ratio & BLEU & chrF++ \\ \midrule
Icelandic & $3.65e-2$ & $1.27e-3$ & $69.24$ & $78.37$ \\
Russian   & $2.01e-2$ & $7.10e-4$ & $68.56$ & $79.15$ \\
Czech     & $1.89e-2$ & $8.15e-4$ & $73.85$ & $82.29$ \\
Chinese   & $1.84e-2$ & $1.64e-4$ & $39.35$ & $58.41$ \\
Ukrainian & $1.63e-2$ & $6.36e-4$ & $71.36$ & $80.85$ \\
Japanese  & $1.03e-2$ & $1.41e-4$ & $54.33$ & $65.44$ \\ \midrule
\textbf{Average}   & $\mathbf{2.01e-2}$ & $\mathbf{6.22e-4}$ & $62.78$ & $74.08$
\end{tabular}%
}
\caption{Language-wise Results for Primary Approach}
\label{tab:my-table}
\end{table}

Table 1 shows that an ensemble of diverse models helps significantly improve the translations, especially in low-resource languages like Icelandic with roughly $3\%$ improvement.

\begin{figure*}[]
    \centering
    \includegraphics[width=1\linewidth]{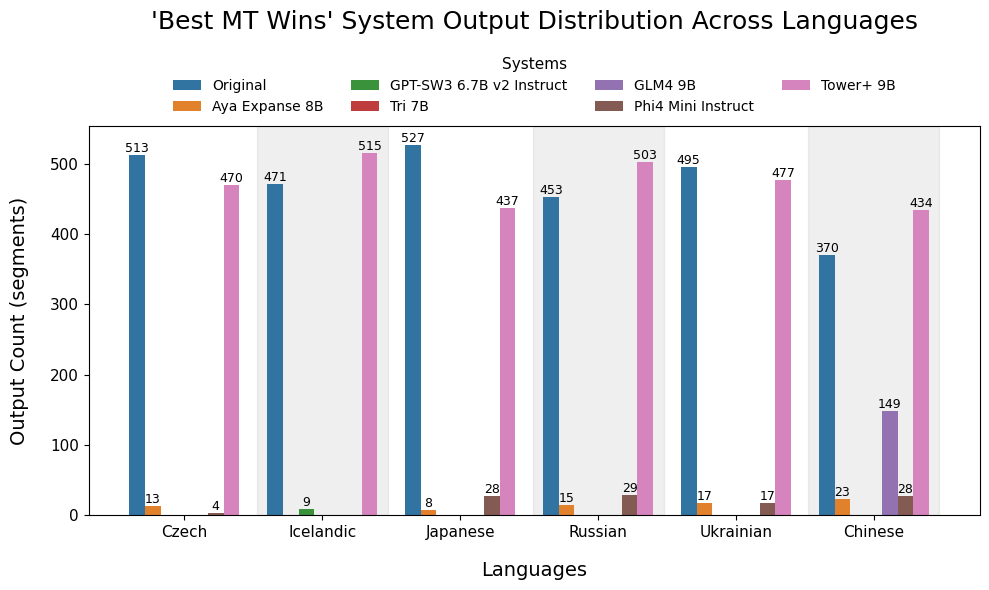}
    \caption{Language-wise count of model responses used in primary approach's final output}
    \label{fig:fig2langwise}
\end{figure*}

As this approach selects the translation with the best QE score, not all systems or model responses contributed equally to the final output. As shown in Table 2, \towerplus and the original translations have contributed the most, while \tri did not contribute at all. Despite \GPTSW, \tri, and \glm trained primarily for use in Nordic and Altaic languages and Chinese, other models provided better translations. Figure~\ref{fig:fig2langwise} shows how different models contributed to the final response in each language.

\begin{table}[h!]
\centering
\begin{tabular}{@{}cc@{}}
\toprule
\textbf{System}          & \textbf{Contribution} \\ \midrule
Tower+ 9B                & 2836                  \\
Original                 & 2829                  \\
GLM4 9B                  & 149                   \\
Phi4 Mini Instruct       & 106                   \\
Aya Expanse 8B           & 76                    \\
GPT-SW3 6.7B v2 Instruct & 9                     \\
Tri 7B                   & 0                     \\ \bottomrule
\end{tabular}
\caption{System-wise Contribution to the Output}
\label{tab:my-table}
\end{table}

\hfill\break

The example in Appendix A is an instance where the direct translation by \towerplus was the best-performing. Interestingly, the corresponding output from the secondary approach is of lower quality. This demonstrates the powerful translation capabilities of the model and the need for better prompt engineering in the secondary approach.

\subsection{Secondary Approach - ``Fill in the Blanks''}

\begin{table}[h!]
\centering
\resizebox{\columnwidth}{!}{%
\begin{tabular}{@{}ccccc@{}}
\toprule
Language & $\Delta$COMET & G2E Ratio & BLEU & chrF++ \\ \midrule
Czech     & $-7.24e-3$ & $-5.80e-7$ & $92.37$ & $95.14$ \\
Russian   & $-8.03e-3$ & $-6.00e-7$ & $92.56$ & $95.31$ \\
Icelandic & $-1.00e-2$ & $-8.20e-7$ & $74.27$ & $82.18$ \\
Chinese   & $-1.30e-2$ & $-4.83e-5$ & $29.52$ & $83.59$ \\
Japanese  & $-1.34e-2$ & $-4.44e-5$ & $25.77$ & $82.59$ \\
Ukrainian & $-1.35e-2$ & $-1.26e-6$ & $91.77$ & $94.43$ \\ \midrule
Average   & $-1.08e-2$ & $-1.59e-5$ & $67.71$ & $88.87$
\end{tabular}%
}
\caption{Language-wise Results for Secondary Approach}
\label{tab:my-table}
\end{table}

Table 3 shows that this approach does not provide much benefit to the overall translation task, with roughly $-0.7$\% to $-1.4$\% quality degradations.

Appendix C is an example where this approach yields a better translation than the original. The original Czech translation by \deepseek has 3 major errors and 1 minor error. This improvement by the model could be owed to the fact that, on top of its multilingual capabilities, \towerplus was fine-tuned on instruction data from different models, including \deepseek. 

There are possible factors why this approach did not perform well: use of bigger and closed-source models in the original MT, use of varied systems and models, inclusion of low-resource languages, correcting only critical + major error spans and ignoring minor error spans in some instances, and more. In the Appendix B example, the original system used Massively Multilingual Neural Machine Translation (MMMT). Although more capable models have been released since then, including \towerplus, output artifacts ("\_\_HEARTBREAK\_\_") and leakage ("Corrected words: [...") affect the translation's quality. This shows that further prompt tuning and output post-processing is needed.

\section{Conclusion and Future Work}

This paper described two translation approaches submitted to the WMT 2025 QE-informed Segment-level Error Correction task. The first approach used QE as a selector among multiple LLM translation outputs, resulting in an overall $\Delta$COMET of $0.0201$. The second approach utilized \towerplus exclusively to replace erroneous words in the MT by masking substrings highlighted in the error spans with a blank token, resulting in $\Delta$COMET of $-0.0108$. Custom prompts were designed to instruct the model in correcting the translation, similar to a fill-in-the-blank task. 

While the first approach showed positive improvements by using QE as a selector, it ultimately depends on the model and system selection. Further exploration of system combinations could yield better performance. The second approach, which performed worse, corrected translations by filling in error segments in the MT. As future work, a two-model system could be explored: a smaller LM to suggest words or phrases for masked tokens via masked language modeling, and a larger LLM to select the most suitable ones to produce a higher-quality translation.

\section*{Limitations}

Due to limited time and compute resources, the overall experimental design favored n-shot prompting with LLMs for their ease of use and availability of pre-trained weights. Additionally, the model selection was guided by convenience and practical factors such as parameter count, recency, and language compatibility or specificity. These limitations also limited the scope for testing and prompt optimization.

Though ``Best MT Wins'' reframes APE as re-translation through QE-based selection, this approach is impractical as it requires generating full hypotheses from large models. While the direct translation capabilities of \towerplus even slightly surpassed the original translation system, the other 5 LLMs used were less effective in comparison, resulting in inefficient use of and wasted time and computational resources.

For the ``Fill in the Blanks'' approach, a lack of proper prompt tuning and post-processing degraded the output quality, indicating that prompt engineering and output handling are important when using LLMs for specific tasks.

More broadly, both approaches rely exclusively on automatic QE metrics such as COMET, which, while effective for shared task evaluation, are primarily trained on English-centric data. The absence of human evaluation limits the ability to validate whether metric-based gains reflect true improvements in translation quality, especially for non-English language pairs.

\section{Acknowledgements}

My sincerest gratitude to Dr Diptesh Kanojia, Archchana Sindhujan and Sourabh Deoghare from the shared task team for their valuable feedback and guidance, encouragement, paper review, and assistance with LaTeX formatting.

\bibliography{custom}

\appendix
\section*{Appendix}  
\addcontentsline{toc}{section}{Appendix}  

\section{Example of Primary Approach}
\label{sec:appendix}
\begin{promptquote}
\textbf{Language}: Chinese \\
\hfill\break
\textbf{Original System}: Wenyiil \\
\hfill\break
\textbf{Original English Text}:\\
A brief note on Play Station controllers to round out the 1st party lineup: I never liked them.\verb|\n|Dual Shock 1-3 felt exactly the same to me with exactly the same flaws: Oversized thumbsticks placed too close together, mushy face/shoulder/trigger buttons that seemed to wear out very quickly, and a D-pad that just felt wrong. Dual Shock 4 fixed a lot of problems, but introduced a bunch of new ones. I got a Dual Sense because some friends had a hard time with asymmetrical thumbsticks, and for some reason local stores were selling them for \$20 less than DS4 controllers.\verb|\n|I don't really use this controller: Face buttons are spaced too far apart, my audio in/output keeps defaulting to the controller when it's connected, the bumpers are weirdly tall, and I keep accidentally pressing the track pad when trying to hit menu/share.\verb|\n|They are fine guest controllers that have come a long way in terms of design, but for me they're a novelty at best.\\
\hfill\break
\textbf{Original Chinese MT}:\\
\begin{CJK}{UTF8}{min}
 关于PlayStation手柄，这里简单提一下，算是对第一方手柄阵容的一个补充：我从来都不喜欢它们。\verb|\n|在我看来，DualShock 1到3代手感完全一样，缺点也如出一辙：摇杆帽太大且位置太近，面板按键、肩键和扳机键手感很肉，而且似乎特别容易磨损，十字键的感觉也总是不对劲。DualShock 4修正了许多问题，但也引出了一堆新毛病。我买了一个DualSense手柄，是因为有些朋友用不惯非对称摇杆，并且出于某种原因，本地商店里它的售价比DS4手柄还要便宜20美元。\verb|\n|我其实不怎么用这个手柄：面板按键间距太开，每次连接手柄时电脑的音频输入/输出都会自动切换过去，肩键的高度很奇怪，而且当我想按菜单/分享键时总会误触到触摸板。\verb|\n|它们作为给客人用的手柄还不错，在设计上也确实有了长足的进步，但对我而言，充其量算是个新奇玩意儿。\\
\end{CJK}
\hfill\break
\textbf{New MT}:\\
\begin{CJK}{UTF8}{min}
 关于 PlayStation 手柄，我想补充一点：我从来不喜欢它们。\verb|\n|Dual Shock 1-3 对我来说感觉完全一样，而且有同样的缺点：拇指摇杆太大，距离太近；脸部/肩部/扳机按钮软绵绵的，似乎很快就会磨损；D 方向键感觉不对。Dual Shock 4 解决了好多问题，但也引入了好多新问题。我买了一个 Dual Sense，因为有些朋友对不对称的拇指摇杆感到不适应，而且不知什么原因，当地商店的价格比 DS4 手柄便宜 20 美元。\verb|\n|我很少用这个手柄：脸部按钮之间的距离太远，连接时我的音频输入/输出总是默认设置为手柄；缓冲器很奇怪地高；当我试图点击菜单/分享时，我总是不小心按下了触控板。\verb|\n|它们是不错的客用手柄，在设计上已经有了很大的改进，但对我来说，它们充其量只是新奇玩意儿。\\
\end{CJK}
\hfill\break
\textbf{Original COMET Score}: 0.7751\\
\hfill\break
\textbf{New COMET Score}: 0.7797\\
\hfill\break
\textbf{Approach 1's Edited COMET Score}: 0.7723
\end{promptquote}

\section{Negative Example for Secondary Approach}
\label{sec:appendix}
\begin{promptquote}
\textbf{Language}: Japanese \\
\hfill\break
\textbf{Original System}: MMMT\\
\hfill\break
\textbf{Original English Text}:\\
Kyle nodded, the corners of his mouth twitching as he fought back a smile. “Is that right? Sorry to let you down. Guess it’s a good thing I hung up the skates when I did, spare a few more fantasy teams the heartbreak.”\verb|\n|Declan opened his mouth, then quickly shut it again, heat creeping up the back of his neck. “I mean…your Cup run with the Kings was insane, though. That game seven OT goal? Unreal.” He scratched the back of his neck. “So… uh, what are you doing at St. Cassian’s? Are you coaching here now? I didn’t see you on the website.”\\
\hfill\break
\textbf{Original Japanese MT}:\\
\begin{CJK}{UTF8}{min}
カイルは頷き、微笑みを抑えながら口角をそらせた。「そうですか。がっかりさせちゃってごめんなさい。急いでスケートを引退しておいてよかったと思うのは、他のファンタジーチームがまた心を痛めることを避けるためでしょう。」\verb|\n|デクランは口を開けてすぐに閉じ、首の後ろに熱さがこみ上げてきた。“いや…キングスとのカップ戦は、ただただすごいね。あの第7戦の延長戦のゴールは…信じられないよ。”彼は首の後ろを掻いた。「それで…あんた、セント・カシアンズでは何をしてるんだ？今はここでコーチをやっているのか？ウェブサイトにはあなたがいる姿が見当たらなかったけど。」
\end{CJK}\\
\hfill\break
\textbf{Modified Japanese MT}:\\
\begin{CJK}{UTF8}{min}
カイルは頷き、微笑みを抑えながら口角をそらせた。「そうですか。がっかりさせちゃってごめんなさい。急いでスケートを引退しておいてよかったと思うのは、他のファンタジーチームがまた\_\_HEARTBREAK\_\_を痛めることを避けるためでしょう。」\verb|\n|デクランは口を開けてすぐに閉じ、首の後ろに熱さがこみ上げてきた。“いや…キングスとのカップ戦は、ただただすごいね。あの第7戦の延長戦のゴールは…信じられないよ。”彼は首の後ろを掻いた。「それで…あんた、セント・カシアンズでは何をしてるんだ？今はここでコーチをやっているのか？ウェブサイトにはあなたがいる姿が見当たらなかったけど。」\\
Corrected words: ['HEARTBREAK', 'カップ戦']
\end{CJK}\\
\hfill\break
\textbf{Original COMET Score}: 0.8183\\
\hfill\break
\textbf{New COMET Score}: 0.7314\\
\hfill\break
\textbf{Error Spans}:\\
$\{'start\_i': 87, 'end\_i': 88, 'severity': 'major'\},$\\
$\{'start\_i': 150, 'end\_i': 154, 'severity': 'major'\}$
\end{promptquote}

\selectlanguage{czech}
\section{Positive Example for Secondary Approach}
\label{sec:appendix}

\begin{promptquote}
\textbf{Language}: Czech\\
\hfill\break
\textbf{Original System}: DeepSeek-v3\\
\hfill\break
\textbf{Original English Text}:
The duchess and her husband, Prince Harry, have to do something and good luck to them, but it's hard not to study the rollout of As Ever for signs of the widening gap between Meghan's self-image and how the rest of the world sees her. ""As ever,"" writes the duchess on Instagram, "means "as it's always been" or some even say "in the same way as always,"" the "some" in this sentence apparently referring to the dictionary definition of a two-word phrase that no one has ever had trouble understanding.\\
\hfill\break
\textbf{Original Czech MT}:
Vévodkyně a její manžel, princ Harry, musí něco dělat, a přejme jim hodně štěstí, ale nelze si nevšimnout, že uvedení značky As Ever odhaluje rostoucí propast mezi Meghaniným sebepojetím a tím, jak ji vnímá zbytek světa. „‚As ever‘,“ píše vévodkyně na Instagramu, „znamená ‚jak to vždy bylo‘, nebo někteří dokonce říkají ‚stejným způsobem jako vždy‘.“ Tito „někteří“ v této větě zřejmě odkazují na slovníkovou definici dvouslovného výrazu, kterému nikdo nikdy nerozuměl.\\
\hfill\break
\textbf{Modified Czech MT}:
Vévodkyně a její manžel, princ Harry, musí něco udělat, a přejme jim hodně štěstí, ale nelze si nevšimnout, že uvedení značky As Ever odhaluje rostoucí propast mezi Meghaniným sebepojetím a tím, jak ji vnímá zbytek světa. „‚As ever‘,“ píše vévodkyně na Instagramu, „znamená ‚jak to vždy bylo‘, nebo někteří dokonce říkají ‚stejným způsobem jako vždy‘.“ Tito „někteří“ v této větě zřejmě odkazují na slovníkovou definici dvouslovného výrazu, kterému nikdo nikdy neměl problém porozumět.\\
\hfill\break
\textbf{Original COMET Score}: 0.8215\\
\hfill\break
\textbf{New COMET Score}: 0.8317\\
\hfill\break
\textbf{Error Spans}:\\
$\{'start\_i': 42, 'end\_i': 53, 'severity': 'major'\},$\\
$\{'start\_i': 174, 'end\_i': 180, 'severity': 'minor'\},$\\
$\{'start\_i': 447, 'end\_i': 467, 'severity': 'major'\},$\\
$\{'start\_i': 468, 'end\_i': 469, 'severity': 'major'\}$\\
\end{promptquote}

\end{document}